\documentclass[journal]{vgtc}                
\pdfoutput=1

\ifpdf
  \pdfoutput=1\relax                   
  \usepackage{graphicx}                
  \DeclareGraphicsExtensions{.pdf,.png,.jpg,.jpeg} 
\else
  \usepackage{graphicx}                
  \DeclareGraphicsExtensions{.eps}     
\fi%

\graphicspath{{./}} 

\usepackage{microtype}                 
\PassOptionsToPackage{warn}{textcomp}  
\usepackage{textcomp}                  
\usepackage{mathptmx}                  
\usepackage{times}                     
\usepackage{cite}                      
\usepackage{tabu}                      
\usepackage{booktabs}                  

\onlineid{0}

\vgtccategory{Research}
\vgtcpapertype{algorithm/technique}

\usepackage{amsmath}
\usepackage{subfig}
\usepackage{svg}

\title{Confluent-Drawing Parallel Coordinates: Web-Based Interactive Visual Analytics of Large Multi-Dimensional Data}

\author{Wenqiang Cui, Girts Strazdins, and Hao Wang, \textit{Member, IEEE}}
\authorfooter{
\item
 Wenqiang Cui is with the Department of ICT and Natural Sciences, Norwegian University of Science and Technology. E-mail: wenqiang.cui@ntnu.no.
\item
 Girts Strazdins is with the Department of ICT and Natural Sciences, Norwegian University of Science and Technology. E-mail: gist@ntnu.no.
\item
 Hao Wang is with the Department of Computer Science, Norwegian University of Science and Technology. E-mail: hawa@ntnu.no.
}

\shortauthortitle{Cui \MakeLowercase{\textit{et al.}}: Confluent-Drawing Parallel Coordinates: Web-Based Interactive Visual Analytics of Large Multi-Dimensional Data}

\abstract{Parallel coordinates plot is one of the most popular and widely used visualization techniques for multi-dimensional data sets. Its main challenges for large-scale data sets are visual clutter and overplotting which hamper the recognition of patterns and trends in the data. In this paper, we propose a confluent drawing approach of parallel coordinates to support the web-based interactive visual analytics of large multi-dimensional data. The proposed method maps multi-dimensional data to node-link diagrams through the data binning-based clustering for each dimension. It uses density-based confluent drawing to visualize clusters and edges to reduce visual clutter and overplotting. Its rendering time is independent of the number of data items. It supports interactive visualization of large data sets without hardware acceleration in a normal web browser. Moreover, we design interactions to control the data binning process with this approach to support interactive visual analytics of large multi-dimensional data sets. Based on the proposed approach, we implement a web-based visual analytics application. The efficiency of the proposed method is examined through experiments on several data sets. The effectiveness of the proposed method is evaluated through a user study, in which two typical tasks of parallel coordinates plot are performed by participants to compare the proposed method with another parallel coordinates bundling technique.  Results show that the proposed method significantly enhances the web-based interactive visual analytics of large multi-dimensional data.%
}


\vgtcinsertpkg

\begin{document}

\maketitle

\section{Introduction}

Parallel coordinates plots (PCP) has become a standard tool for visualizing multi-dimensional data~\cite{Hei2013}. It has been incorporated into several visual analytics applications for exploring multi-dimensional data sets, such as iPCA~\cite{Jeo2019} and iVisClassifier~\cite{Cho2010}. In PCP, axes, that corresponding to dimensions of the original data, are aligned parallel to each other and data points are mapped to lines intersecting the axes at the respective value. The embedding of an arbitrary number of parallel axes into the plane allows the simultaneous display of many dimensions, providing a good overview of the data, which reveals patterns and trends in the data. However, it often creates visual clutter, especially for large-scale data sets. This hampers the recognition of patterns and trends in the data. In addition, another related issue of PCP for large-scale data sets is overplotting, in which lines are plotted on top of each other. This hides information in the data.

Edge bundling is a widely used approach to reduce visual clutter in PCP. The data will be aggregated first with data clustering algorithms, then rendered in an illustrative fashion using different forms of edge bundling. However, existing data clustering and edge-bundling techniques used in PCP are not scalable well for large-scale data. Their rendering performance depends on the number of data items. For millions of data items, they have been too slow for interactive use without any hardware acceleration, for example, in web-based visual analytics, which hinders the integration of human judgment into the data analysis process.  Furthermore, they do not support to modify both the number and the size of the clusters generated by automatic algorithms, which hinders their use in visual analytics. Moreover, they do not address the issue of overplotting.

Web-based interactive visual analytics is a widely used in various application domains. Many researches have been conducted to explore web-based interactive visual analytics for various types of data, such as miRTarVis+~\cite{Seh2017}, GraphVis~\cite{Ahm2015}, and a web-based visual analytics system for real estate data~\cite{Sun2013}. However, existing web-based visual analytics methods are not scalable well for large data. The time to process data and render visualizations hinders their interactive use in web browsers when the size of data sets is scaled up.

In this paper, we propose a confluent-drawing-based PCP approach for web-based visual analytics of large multi-dimensional data sets. It maps multi-dimensional data to node-link diagrams to reduce unambiguous visual clutter through a data binning-based clustering for each dimension. It visualizes the data in density-based confluent drawing to address overplotting. It is scalable for large data sets and supports interactive use in the web-based visual analytics. Based on the proposed confluent-drawing PCP, we implement a web-based visual analytics application for large multi-dimensional data sets. The experiments on several data sets are conducted to demonstrate the scalability of the proposed approach. In addition, a user study is conducted to compare its effectiveness against another parallel coordinates bundling technique. Results show that it enhances the web-based interactive visual analytics of large multi-dimensional data in the following ways:

\begin{enumerate}
	\item To the best of our knowledge, our confluent-drawing PCP is the first PCP approach that supports interactive use on large data sets (e.g. millions of data items) in web browsers without hardware-accelerated rendering and big data infrastructure-based data processing.
	\item It reduces visual clutter and overplotting through data binning-based clustering for each dimension and density-based confluent drawing, which eliminates the ambiguity in other bundling techniques that bundle edges by spatial proximity. 
	\item The rendering time of our method is independent of the number of data items. By applying confluent drawing on PCP, it merges the "links" before rendering them and uses their densities to render their widths to reveal the data distributions between axes.
	\item Our method can be used to detect anomalies that have extremely small proportions in large multi-dimensional data sets.
	\item To support visual analytics of large multi-dimensional data, our method provides interactions to directly modify both the number and the size of the clusters in the visualization.
\end{enumerate}

The paper is organized as follows:~\autoref{section:related work} reviews related work. \autoref{section:method} presents the method. \autoref{section:experiment} conducts the experiments and user study, and discusses the result. \autoref{section:future} draws conclusions.

\section{Related Work}\label{section:related work}

Since parallel coordinates were introduced in the seminal work of Inselberg~\cite{Ins1985,Ins1987}, and later extended by Wegman~\cite{Weg1990} as a visualization tool, it has become a common and widely used visualization technique for visualizing multi-dimensional data. For large-scale data sets, visual clutter and overplotting are often cited as major challenges for parallel coordinates~\cite{Hei2013}. Use data clustering algorithms to group data items and display the clusters in extensions of PCP with techniques such as edge bundling is a usual way to reduce visual clutter. Fua et al.~\cite{Fua1999} develop a multiresolutional view of the data via hierarchical clustering, and use a variation on parallel coordinates to convey aggregation information for the resulting clusters. To visualize a large number of data items without hiding the inherent structure they constitute, Johansson et al. \cite{Joh2005} combine clusters and high-precision textures in PCP to represent data, in which specific transfer functions that operate on the high-precision textures are used to highlight different aspects of the cluster characteristics.

The edge-bundling method that uses cubic B-splines to show adjacency relations atop different tree visualization methods was proposed for the first time by Holten et al. \cite{Hol2006}.  It has been adapted to PCP based on data clustering by McDonnell and Mueller~\cite{Mcd2008}. With edge bundling, polylines in PCP become polycurves, which are bent between two axes towards centroid of the corresponding clusters. While this approach reduces visual clutter by freeing up screen space, the resulting polycurves are still $C^{0}$-continuous, which makes them difficult to be visually traced. To avoid this, Heinrich et al.~\cite{Hei2011} use a $C^{1}$-continuous bundling. Furthermore, Zhou et al.~\cite{Zho2008} present a bundling method that creates visual clusters without the prerequisite of data clustering.

Since edge bundling was introduced, multiple approaches have been proposed for diverse purposes, such as ink saving, reducing ambiguity or revealing information hidden in the visualization~\cite{Lhu2017}.  To visualize non-planar graphs in a planar way and reduce ambiguity, Dickerson et al.~\cite{Dic2003} propose an unambiguous edge bundling called confluent drawing, which allows groups of edges to be merged together and drawn as "tracks" to facilitate tasks where the user needs to follow lines. Luo et al.~\cite{Luo2008}  introduce a model for curve bundles in PCP to reduce visual clutter and display inherent structure in the data by vertically redistributing and separating the clusters. Palmas et al.~\cite{Pal2014} present an edge-bundling layout for PCP using density-based clustering for each dimension independently so that the clustering is directly related to the shown dimensions in every part of the plot. Do Amor Divino Lima et al.~\cite{Do2018} propose an edge-bundling technique to visually encode the clusters information of each dimension, such as variance, means, and quartiles, into the curvature of lines.

Existing work on edge bundling concentrates on combining various clustering algorithms and different forms of edge bundling with parallel coordinates. There is a lack of works in the literature about edge bundling and PCP that focus on supporting their interactive use for visual analytics of large-scale data sets, especially, without hardware accelerations. The long computation and rendering time makes them too slow for interactive use in the visual analytics process. For example, in Palmas et al.'s~\cite{Pal2014} method, to reduce the computation time for the clustering, they sample the parameters 2000 times between 25\% and 1\% of the data range instead of computing it on the whole data range. However, this precomputation still needs almost one minute for $10^{5}$ data points on desktop hardware in a single computing thread. In addition, its rendering performance depends on the number of data items, which hinders the interactive exploration for visual analytics when the number of data items is scaled up. In contrast, our method takes advantage of human expertise by supporting directly modifying both the number and the size of clusters of millions of data items in the visualization. Its rendering performance is independent of the number of data items.

Density representation is another usual way to address overplotting in PCP~\cite{Mil1991,Weg1997}. Artero et al.~\cite{Art2004} highlight significant relationships between axes in a PCP by constructing a density plot based on density information computed from the data set. Heinrich et al.~\cite{Hei2009} derive a mathematical density model to generate a continuous parallel coordinates plot. Kosara et al.~\cite{Kos2006} introduce a method called parallel sets that uses parallelograms to visualize categorical data in PCP, where the thickness of the parallelogram reveals the number of data points that are in both of the two connected categories. To reveal the internal structure of clusters and the data distribution between axes, we follow~\cite{Kos2006} in using the densities of bundles to render their widths. It is effective even with a purely geometry-based visualization, which leaves the color and opacity perceptual channels free for visualizing other information.

The ever-increasing amount of data poses a significant challenge for interactive visual analytics, especially, for the web-based interactive visual analytics. Sansen et al.~\cite{San2017} present a system for visual exploration of large multi-dimensional data with parallel coordinates based on a big data infrastructure, which employs a hybrid computing method to accommodate pre-computing time and hardware-accelerated rendering to reduce rendering time. However, their method is hard to use in the web-based interactive visual analytics without the big data infrastructure and hardware-accelerated rendering.

\section{Method} \label{section:method}

With our confluent-drawing PCP technique, the data is clustered through a data binning-based clustering method for each dimension (\autoref{section:dataclustering}). After clustering, it first maps the multi-dimensional data into a node-link diagram and then uses density-based confluent drawing to visualize the data within B\'ezier splines in PCP (\autoref{section:edgemerging}). In addition, it provides interactions to support the web-based interactive visual analytics (\autoref{section:interaction}).

\subsection{Data Binning-Based Clustering} \label{section:dataclustering}

Data binning is the process to group a number of more or less continuous values into a smaller number of given data intervals (also called "bins"), which transforms numerical variables into categorical counterparts. We use data binning to cluster the data points for each dimension (continuous variable) with the following two considerations:
\begin{enumerate}
	\item In a parallel coordinates plot, the data points in each axis are ordered and one data point will be grouped into only one cluster.
	\item For a complex data set, to take advantage of human reasoning, perception, and cognition, users may prefer to specify the clusters based on their own knowledge than an automated algorithm.
\end{enumerate}

With the first consideration, for a specified cluster number $k$, the clusters in one axis are also ordered and there is no overlap between clusters, which is illustrated in \autoref{fig:clustersinaxis}. In \autoref{fig:clustersinaxis}, we use a center and a radius to define a cluster in one axis, where the gray rectangles represent clusters, the blue dots represent the centers of the clusters, and the green lines represent the radius of the clusters. With this definition, $k$ clusters in one axis  for a given data set $D$ are expressed as follows:
\begin{figure}[tb]
	\centering
	\includegraphics[width=0.9\linewidth]{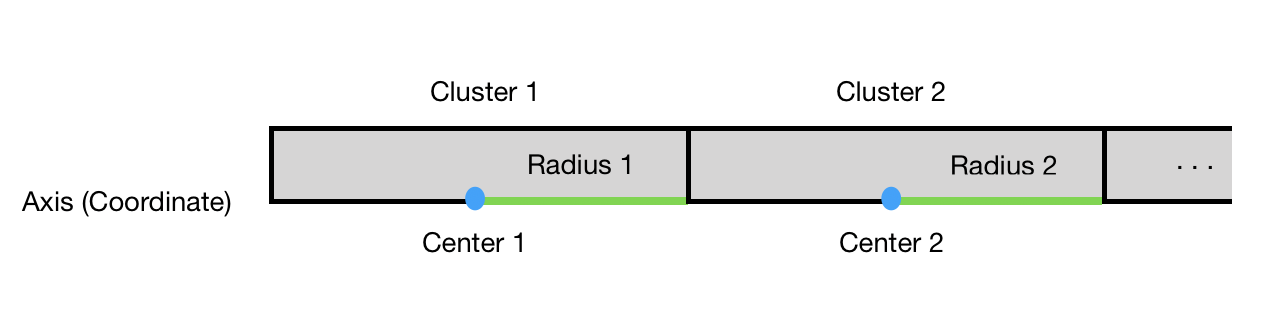}
	\caption{\label{fig:clustersinaxis}
		\textit{Clusters in one (horizontal) axis of a PCP.}}
\end{figure} 
\[
\left\lbrace (C_{1},R_{1}), (C_{2},R_{3}), . . ., (C_{k},R_{k})\right\rbrace 
\]
which fulfills the following constrains:
\[
C_{i}<C_{i+1}, R_{i} \leq C_{i+1} - R_{i+1} - C_{i}, \sum_{i=1}^{k} R_{i} \leq (d_{max}-d_{min})/2
\]
where $C_{i}$ is the center of the $i$-th cluster, $R_{i}$ is the radius of the $i$-th cluster, $d_{max}$ is the maxima of the data set $D$, and $d_{min}$ is the minima of the data set $D$. 

To use this data binning method in our confluent-drawing PCP, we first specifically select $k$ centers which evenly group the data points into $k$ clusters (bins) for each dimension. Accordingly, in each dimension, the $k$ ordered clusters with the same radius can be obtained without iterations through the following computation:
\[
C_{i}=d_{max}-R\times(1+2\times i), R = \frac{(d_{max}-d_{min})}{2 \times k}
\] 

This computation only relies on the maxima, minima of the data points and the number of the clusters for each dimension. This will eliminate most of the computational time required for other clustering methods used in edge-bundling PCP, such as Gaussian kernel density estimation [19] and DBSCAN [4], especially for large data sets.

Furthermore, with the second consideration, to take advantage of human perception and reasoning abilities, our method allows users to directly modify the number and size of clusters in the visualization with designed interactions, which is demonstrated in \autoref{section:interaction}. 

Categorical variables are not clustered using the above method. Instead, we treat each category as a cluster. Using this method, the time of computing arbitrary clusters for a large data set could be ignored with the known maxima and minima. 

\subsection{Density-Based Confluent Drawing on PCP} \label{section:edgemerging} 

\begin{figure}[tb]
	\centering
	\includegraphics[width=0.9\linewidth]{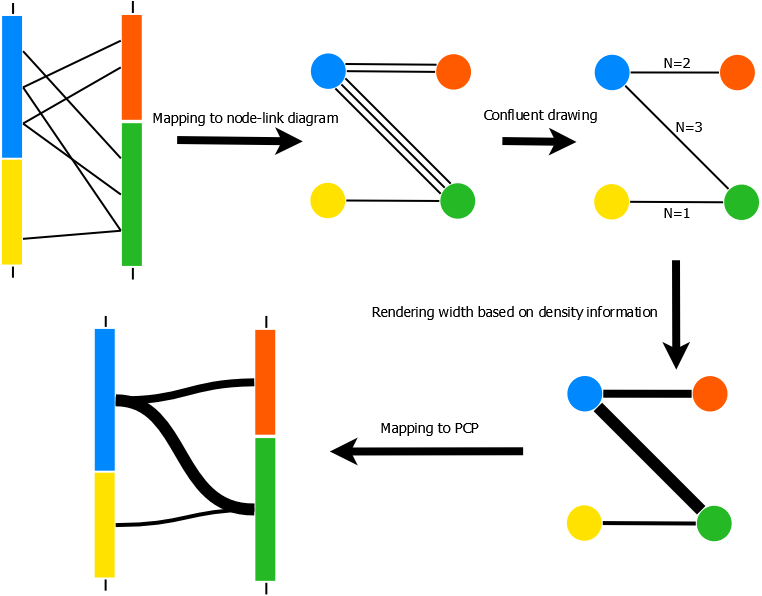}
	\caption{The complete process of confluent drawing on PCP.}
	\label{fig:vis_process}
\end{figure}

Based on the result of the clustering process, our method maps multi-dimensional data into a node-link diagram, where the clusters are mapped as the "nodes" and the data points that are in both of the two connected clusters are mapped as the "links". Then, confluent drawing is applied on the node-link diagram to merge the links. Finally, the node-link diagram with confluent drawing is visualized on PCP, where the merged links are rendered as B\'ezier splines to guarantee $C^{1}$ continuous across axes. During the confluent drawing process, the number of the links between two nodes is counted and used to calculate the density of the merged link. When visualizing the node-link diagram on PCP with confluent drawing, the densities of merged links are used to render their widths to reveal the data distributions between axes. This complete process is shown in~\autoref{fig:vis_process}.

As the process of mapping clusters to the node-link diagram shown in~\autoref{fig:vis_process}, the clusters in each axis are mapped as nodes, and the data points that belong to a pair of clusters are mapped as links connecting two nodes. To group the data points within two neighbor axes into cluster pairs, we use the following algorithm:
\begin{align*}
&	for \; all  \;  data \; points \; in \; two \; neighbor \; axes \; :\\
&	\quad if \; a \in C_{A}^{i} \; and \; b \in C_{B}^{j}\\
&	\quad then \; (a,b) \in P(C_{A}^{i}, C_{B}^{j})
\end{align*}
where, $A$ and $B$ are the names of the axes, $a$ and $b$ are the data points in $A$ and $B$, $C_{A}^{i}$ is the $i$-th cluster in the axis $A$, $C_{B}^{j}$ is the $j$-th cluster in the axis $B$, and $P(C_{A}^{i}, C_{B}^{j})$ is the pair of two clusters. 

Confluent drawing is a technique for bundling links in node-link diagrams. It coalesce groups of lines into common paths or bundles based on network connectivity to reduce edge clutter in node-link diagrams~\cite{Dic2003, Bac2017}. This eliminates the ambiguity in other bundling techniques that bundle edges by spatial proximity. Our method uses confluent drawing to merge the links that belong to cluster pairs in the node-link diagram, which is shown in \autoref{fig:vis_process} as confluent drawing process.

In addition, our method computes the density for each bundle as:
\[
D_{i,j} = \frac{N(C_{A}^{i}, C_{B}^{j})}{\sum_{i=1}^{i}\sum_{j=1}^{j}N(C_{A}^{i}, C_{B}^{j})}
\]
where, $D_{i,j}$ is the density of one bundle, $N(C_{A}^{i}, C_{B}^{j})$ is the number of data points (links) that belong to the cluster pair $P(C_{A}^{i}, C_{B}^{j})$.

The density of a bundle is used to render its width to reveal the data distribution between axes, which is shown in \autoref{fig:vis_process} as the process of rendering bundles' widths. The width of each bundle is computed as:
\[ 
W_{i,j} = D_{i,j} \times W_{max}
\]
where, $W_{i,j}$ is the width of one bundle, and $W_{max}$ is the max bundle width in the whole visualization. $W_{max}$ is configurable for different data sets. Therefore, for two neighbor axes, the width of a bundle represents the proportion of data items that are in both of the two connected clusters. For each pair of clusters, the larger the proportion of data items is, the wider the bundle is. The smaller the proportion of data items is, the narrower the bundle is. This allows our method to detect anomalies (rare items or observations which raise suspicions by differing significantly from the majority of the data~\cite{Liu2009}) in a large multi-dimensional data set through visualizing the extremely narrow bundles in a different style, for example, a dashed line, which is demonstrated in~\autoref{section:experiment_result}.

Finally, the node-link diagram with confluent drawing is visualized on PCP, which is shown in \autoref{fig:vis_process} as the process of mapping to PCP. In classic PCP, data points are drawn as polylines, which loses visual continuation across axes. To guarantee the $C^{1}$ continuous across axes, our method displays the bundles as B\'ezier splines, which is a common method in PCP.

\subsection{Web-Based Interactive Visual Analytics} \label{section:interaction}
In addition to the typical interactions of classic PCP, such as changing the order of axes, our method provides interactions that are specifically designed for its data binning-based clustering process to support interactive visual analytics. To support these interactions, it adds control points between clusters in each axis, which is shown in \autoref{fig:control_points}. The interactions are described as follows:

\begin{figure}[!t]
	\centering
	\includegraphics[width=0.9\linewidth]{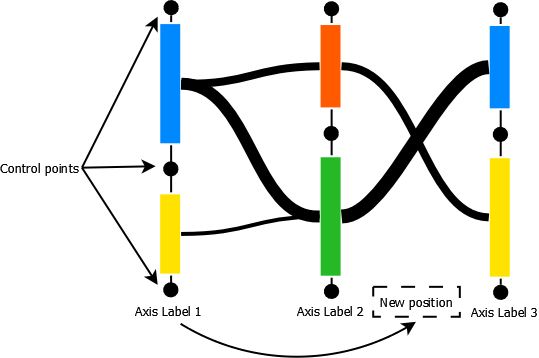}
	\caption{\label{fig:control_points}
		The interactions for interactive visual analytics.}
\end{figure} 
\begin{description}
	\item [Change the order of the axes.] One-click on the label of an axis to select the axis and drag it to a new position. Then the order of the axes will be rearranged and the visualization will be re-rendered.
	\item [Modify the size of clusters.] One-click on a control point and drag it along the axis to a new position on the axis. Then the size of the clusters that are adjacent to it will be modified and the visualization will be re-rendered.
	\item[Add a cluster.] Double-click on a cluster. Then a control point will be added at the clicked position to split the original cluster and the visualization will be re-rendered based on the new clustering results.
	\item[Delete a cluster.] Double-click on a control point. Then the control point will be deleted, the two clusters that are adjacent to it will be merged into a bigger cluster, and the visualization will be re-rendered based on the new clustering result.
\end{description}

\begin{figure}[tb]
	\centering
	\includegraphics[width=0.9\linewidth]{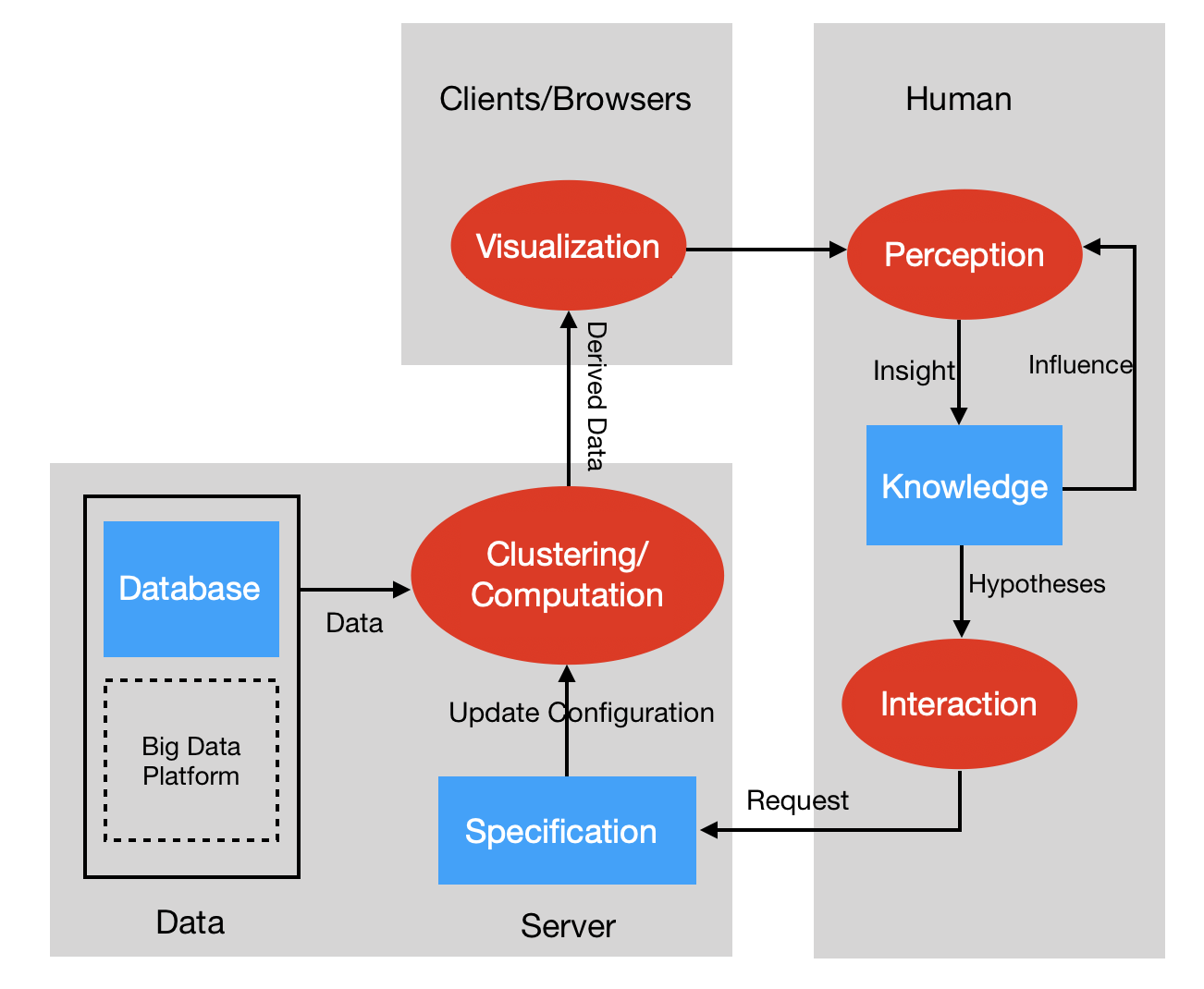}
	\caption{\label{fig:va_process}
		The visual analytics and knowledge generation process of the visual analytics application.}
\end{figure}

Without hardware-accelerated rendering and big data infrastructure-based data processing, the scalability of our method supports its interactive use in the web-based visual analytics of large multi-dimensional data. Its scalability is demonstrated in \autoref{section:experiment_result}. We implement our method in a web-based visual analytics application using Java with Spring MVC on the server side and Data-Driven Documents (D3.js) on the client side. It supports visual analytics by combining data clustering, visualization and human judgment through user interactions. Its architecture and knowledge generation model of visual analytics are shown in \autoref{fig:va_process}. 

\begin{figure}[tb]
	\centering
		\includegraphics[width=0.9\linewidth]{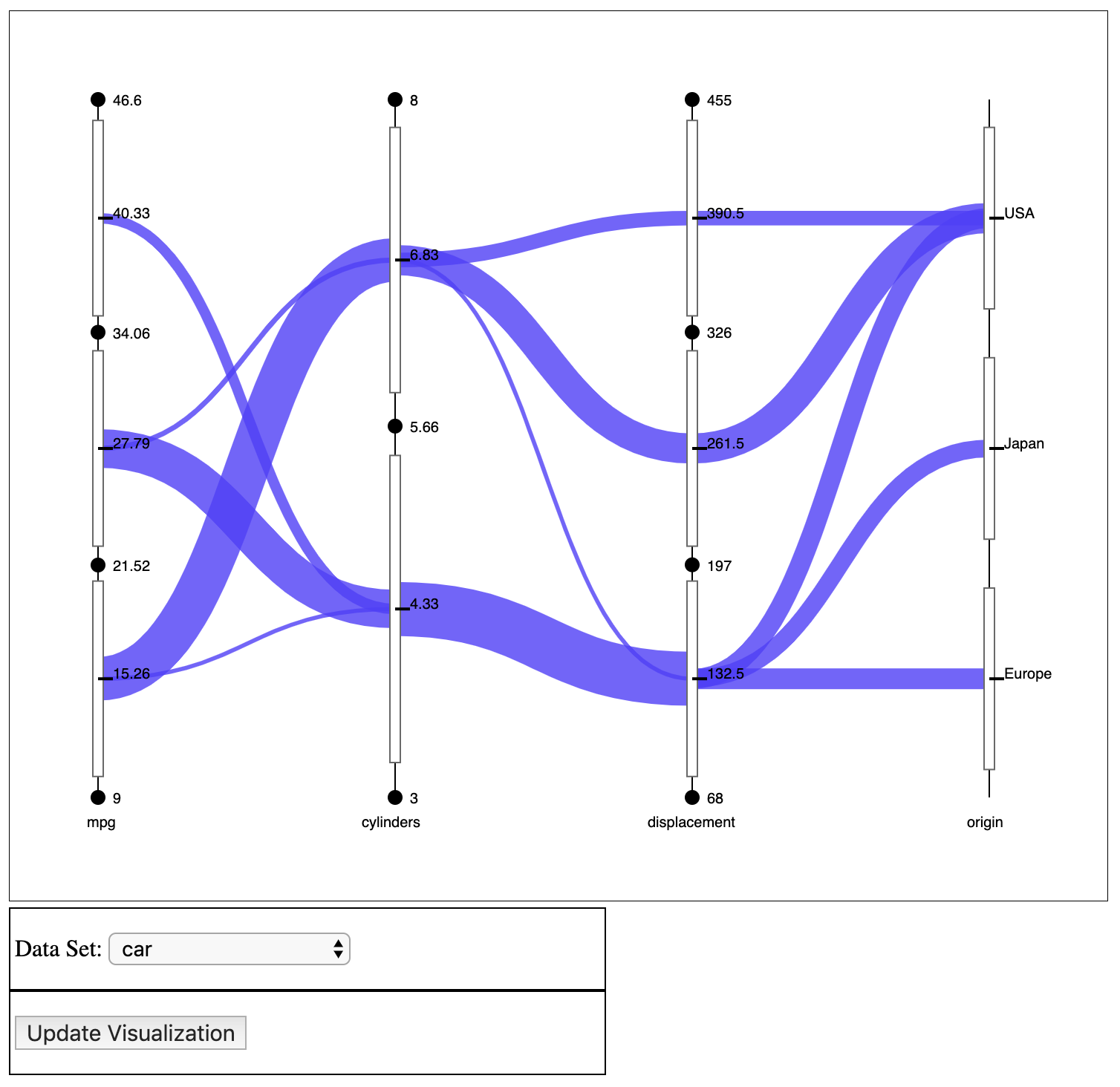}
	\caption{\label{fig:system} 
		Confluent-drawing PCP on four dimensions from the cars data set.}
\end{figure}

\autoref{fig:system} is the screen-shot of the web-based visual analytics application. It shows the confluent-drawing PCP on four dimensions from the cars data set~\cite{Cars}. In this PCP layout, the total density of all bundles between each two neighbor axes is 1, therefore, the widths of the bundles reflect the distribution of the data points within the axes. 

Please see the video in the supplemental material for an impression of our application and the interactions mentioned above.

\section{Experimental Results and User Study} \label{section:experiment}
In this section, we evaluate the scalability and the effectiveness of the proposed approach through experiments and a user study on the office occupancy detection data set (five attributes and 20560 data points for each attribute)~\cite{Can2016} and the cars data set (seven attributes and 392 data points for each attribute)~\cite{Cars}. The office occupancy detection data set uses the measurement data of temperature, humidity, light, and CO2 to detect the occupancy of an office room. The cars data set consists of data on cylinders, horsepower, weight, etc. for cars. In addition, to examine the scalability of our method on large data sets, we also synthesized several larger data sets based on the office occupancy detection data set. All the experiments were performed on a MacBook Pro desktop with 3.1 GHz Intel Core i5 CPU, 8GB Memory and Intel Iris Plus 650 Graphics Processing Unit (GPU) without hardware-accelerated computing and rendering. 

\subsection{Experimental Results}\label{section:experiment_result}
Two experiments are performed to evaluate the scalability of the proposed method. First, a run-time analysis of the data binning-based clustering method was performed. The clustering process includes calculation of the clusters for each dimension and the densities of bundles. The result of the run-time analysis is shown in \autoref{tab:run_time_clustering}, which shows the computation time of our method is linearly dependent on the number of dimensions and the number of data points, and independent on the number of clusters. Furthermore, the computation time is much shorter compared with other clustering algorithms used in PCP. For example, Palmas et al. \cite{Pal2014} present a density-based clustering for each dimension independently in edge-bundling PCP, which takes about one minute for clustering a one-dimensional data set with $10^{5}$ data items. In contrast, our algorithm takes 0.023 seconds for clustering a four-dimensional data set with $10^{5}$ data items, which allows the interactive use of it in visual analytics.

\begin{table}[tb]
	\caption{Run-time analysis of the data binning-based clustering.}
	\label{tab:run_time_clustering}
	\scriptsize%
	\centering%
	\begin{tabu}{%
			*{3}{c}%
			l%
		}
		\toprule
	 	\rotatebox{90}{Number of} \rotatebox{90}{dimensions}  & \rotatebox{90}{Number of} \rotatebox{90}{data points} & \rotatebox{90}{Number of} \rotatebox{90}{clusters} & \rotatebox{90}{Run-time} \rotatebox{90}{(in seconds)}   \\
		\midrule
		2 & $10^4$ & 3  & 0.002 \\
		2 & $10^4$ & 4  & 0.002 \\
		2 & $10^5$ & 3 & 0.017 \\
		2 & $10^5$ & 4 & 0.018 \\
		3 & $10^4$ & 3  & 0.0045 \\
		3 & $10^4$ & 4  & 0.004 \\
		3 & $10^5$ & 3  & 0.022 \\
		3 & $10^5$ & 4  & 0.022 \\
		4 & $10^5$ &  3 & 0.023 \\
		4 & $10^5$ &  4 & 0.03 \\
		4 & $10^6$ & 3  & 0.12 \\
		4 & $10^6$ & 4  & 0.123 \\
		\bottomrule
	\end{tabu}%
\end{table}

\begin{figure*}[tb]
	\centering
	\subfloat[The office occupancy data set visualized in the classic PCP\label{room1}]{%
		\includegraphics[width=0.48\linewidth]{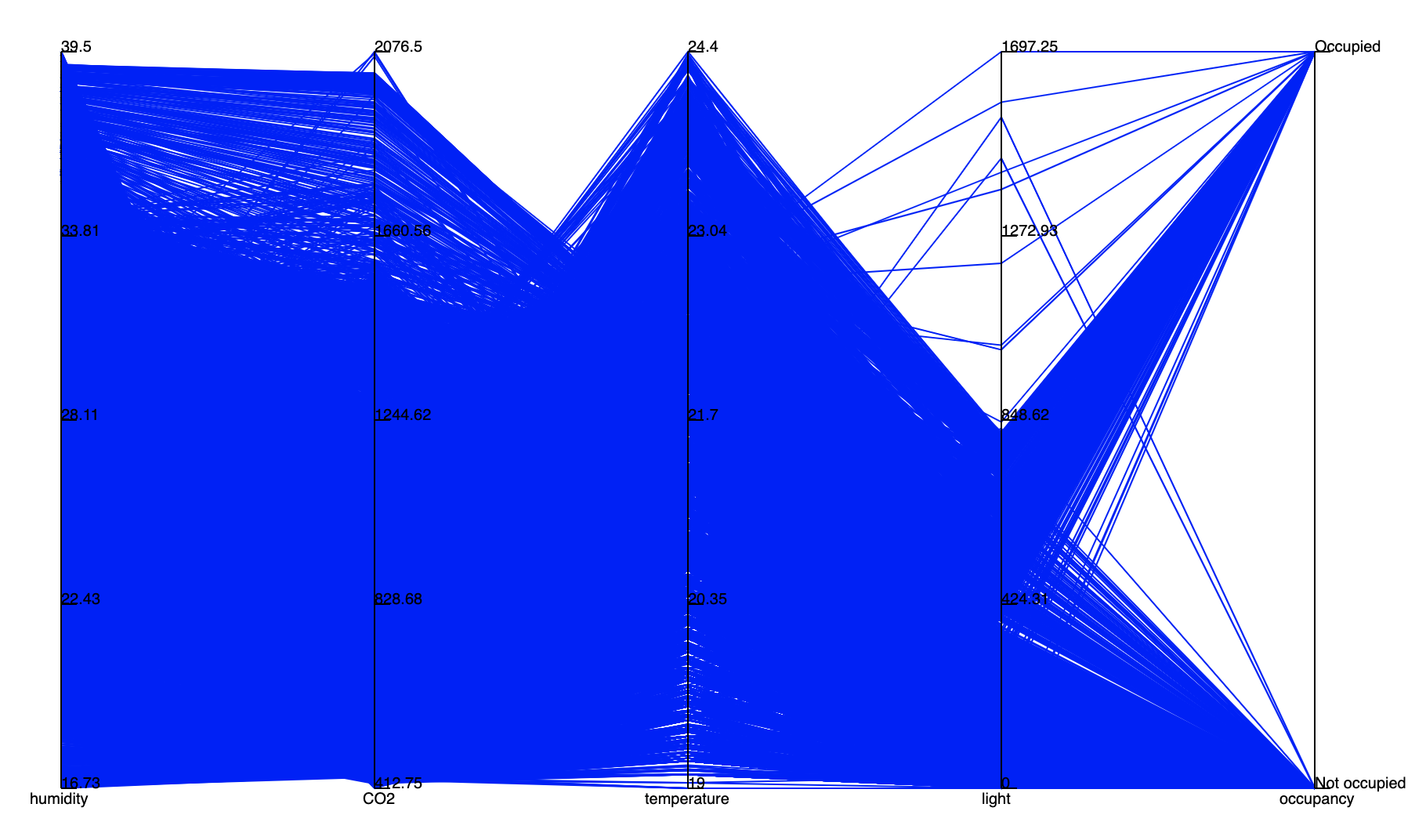}}
	\hfill
	\subfloat[The office occupancy data set visualized in do Amor Divino Lima et al.~\cite{Do2018}'s edge-bundling PCP.\label{room2}]{%
		\includegraphics[width=0.48\linewidth]{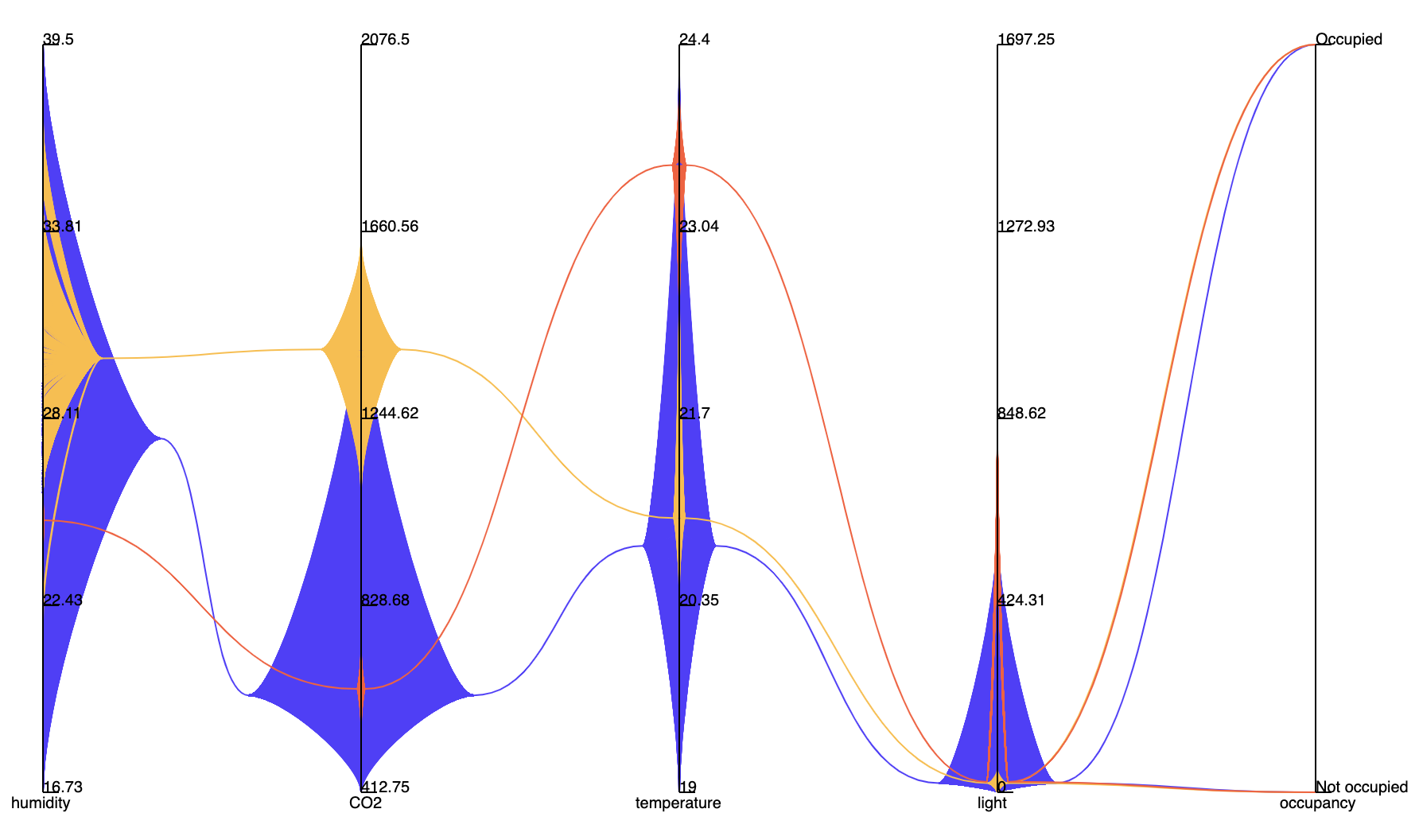}}
	\hfill
	\subfloat[The office occupancy data set visualized in our confluent-drawing PCP without anomalies (dashed lines)\label{room3_na}]{%
		\includegraphics[width=0.48\linewidth]{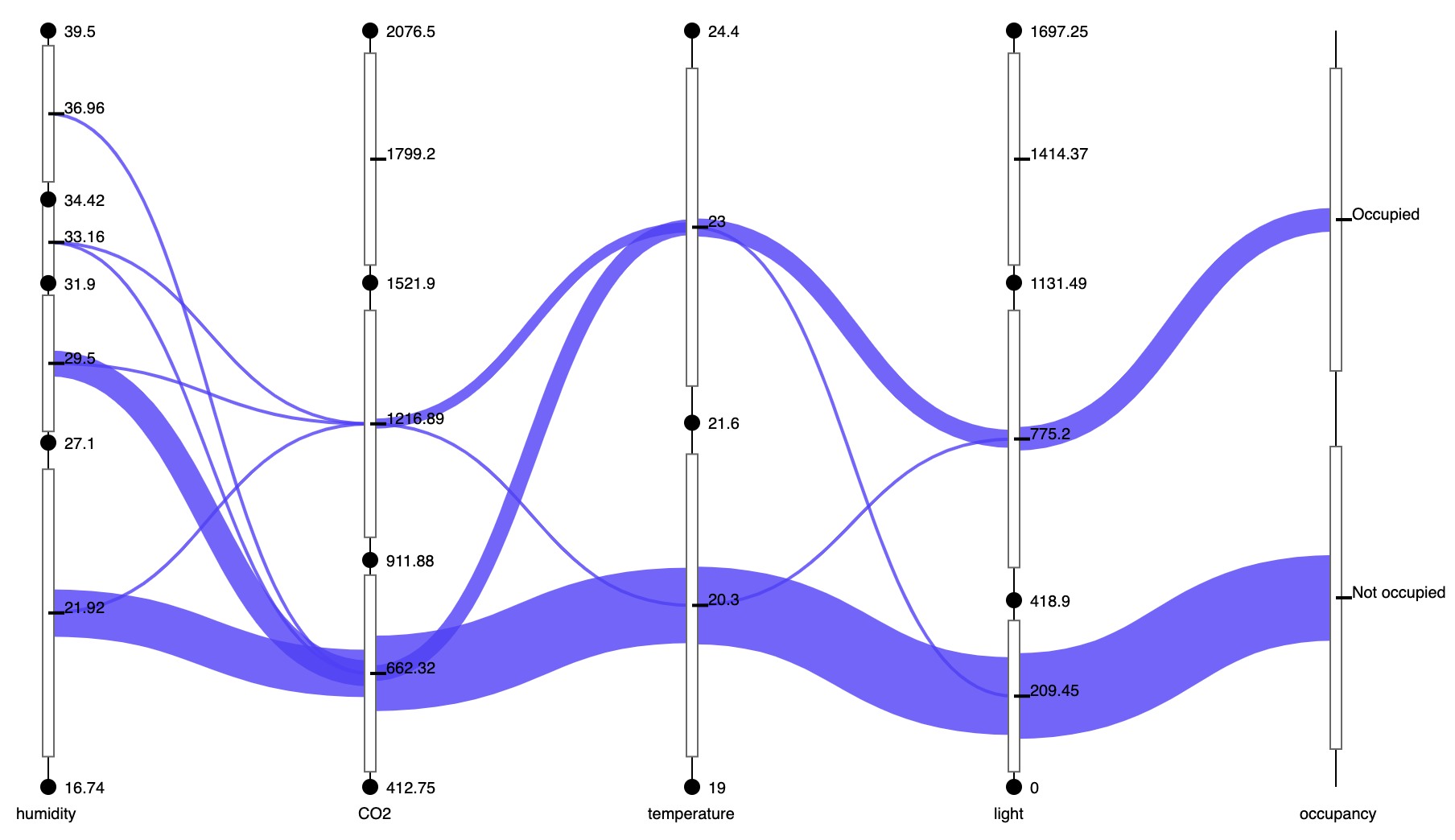}}
	\hfill
	\subfloat[The office occupancy data set visualized in our confluent-drawing PCP with anomalies (dashed lines)\label{room3}]{%
		\includegraphics[width=0.48\linewidth]{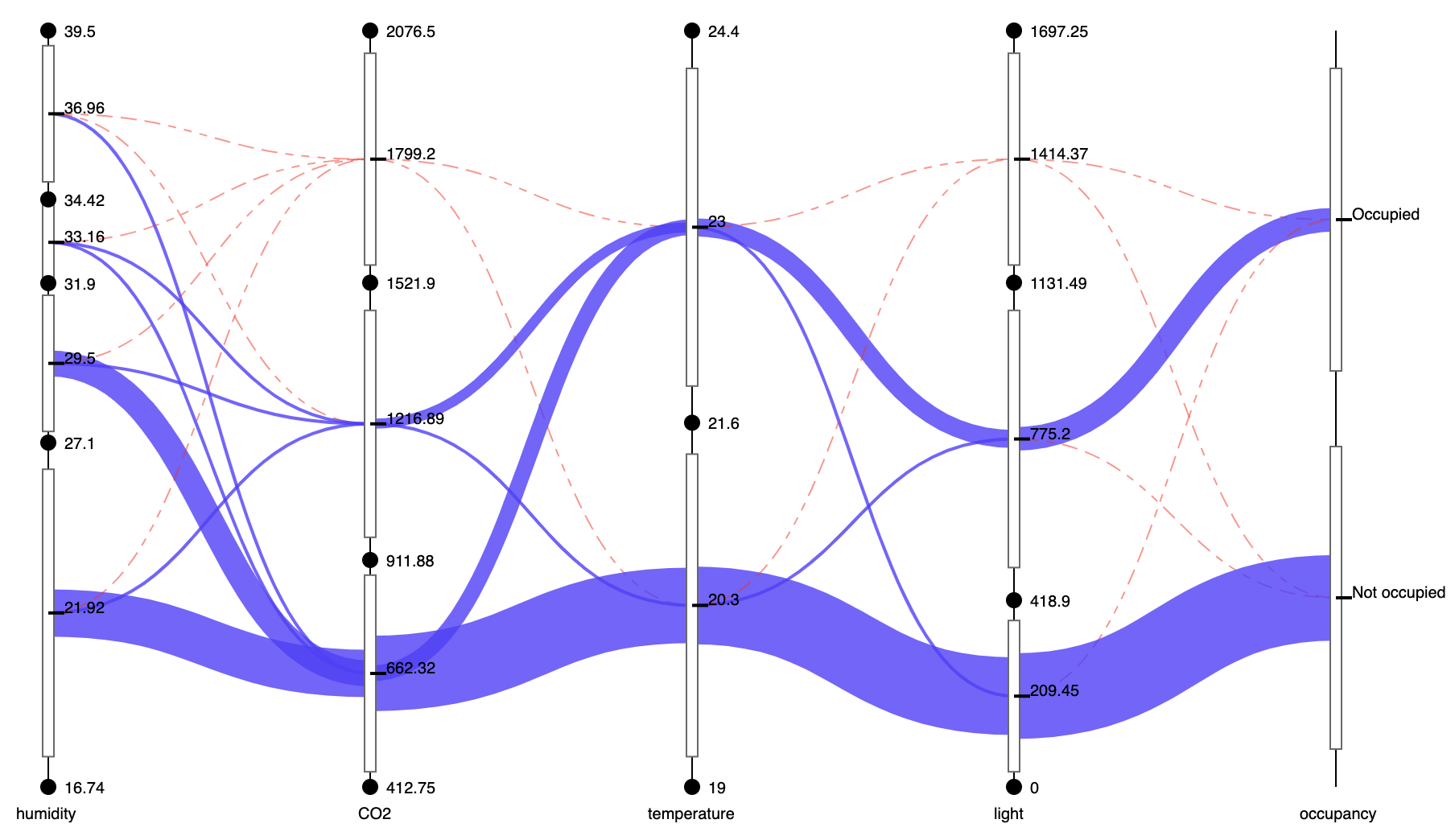}}
	\hfill
	\caption{The comparison of different PCP layouts with the office occupancy data set.}
	\label{fig:pcp_comparison} 
\end{figure*}

Next, we examined the efficiency of our method by comparing its rendering time with both classic PCP and do Amor Divino Lima et al.'s edge-bundling PCP~\cite{Do2018} using the office occupancy detection data set. All methods are implemented in D3.js in Chrome (version: 73.0.3683.103, 64-bit). \autoref{fig:pcp_comparison} shows the comparison of these three different PCP layouts. Particularly, \autoref{room3} is an example of anomaly detection by using our method. In~\autoref{room3}, the extremely narrow bundles are visualized in dashed lines to represent anomalous data items that has extremely small proportions between each two neighbor axes.

To make the comparison fair, only time for rendering the data items is included. The time for rendering the axes, labels and stickers are not included - this part is static and requires a constant rendering time regardless of the number of data items. The result is shown in \autoref{tab:run_time_rendering}, in which the rendering time is counted in seconds.

\begin{table}[tb]
	\caption{Comparison of rendering time of different PCP layouts.}
	\label{tab:run_time_rendering}
	\scriptsize%
	\centering%
	\begin{tabu}{%
			*{2}{c}%
			*{2}{l}%
		}
		\toprule
		\rotatebox{90}{Number of} \rotatebox{90}{data points}  & \rotatebox{90}{Rendering time of} \rotatebox{90}{our method} & \rotatebox{90}{Rendering time of} \rotatebox{90}{classic PCP}  & \rotatebox{90}{Rendering time of} \rotatebox{90}{edge-bundling PCP}   \\
		\midrule
		$10^3$ & 0.00115  & 0.0261 & 0.0735 \\
		$10^4$ & 0.00136  & 0.1505 & 0.92712 \\
		$5\times10^4$ & 0.00137 & 0.65785 & 1.53715 \\
		$10^5$ & 0.00117 & 1.44726 &26.3704 \\
		$10^6$ & 0.00136  & N/A & N/A \\
		\bottomrule
	\end{tabu}%
\end{table}

According to \autoref{tab:run_time_rendering}, the classic PCP and do Amor Divino Lima et al.'s edge-bundling PCP take more than 1.4 and 26 seconds for visualizing $10^5$ data items and cause a crash in the browser with $10^6$ data items. In contrast, our method takes about 0.001 seconds for rendering the data sets. The rendering time of our method is much shorter compared to that of classic PCP and do Amor Divino Lima et al.'s edge-bundling PCP. This is because our method maps the data into the node-link diagram and merges data points before rendering them by applying confluent drawing on the data. For given two axes, the maximum of bundles between them only depends on the number of clusters in each axis, which is much smaller and constant when comparing with the number of data points in a large data set. The rendering performance of our method is independent on the number of data items, which makes our method is scalable for large data sets.

\subsection{User Study}
\begin{figure} [tb]
	\centering
	\subfloat[The car data set visualized in do Amor Divino Lima et al.~\cite{Do2018}'s edge-bundling PCP.\label{correlationEB}]{%
		\includegraphics[width=0.9\linewidth]{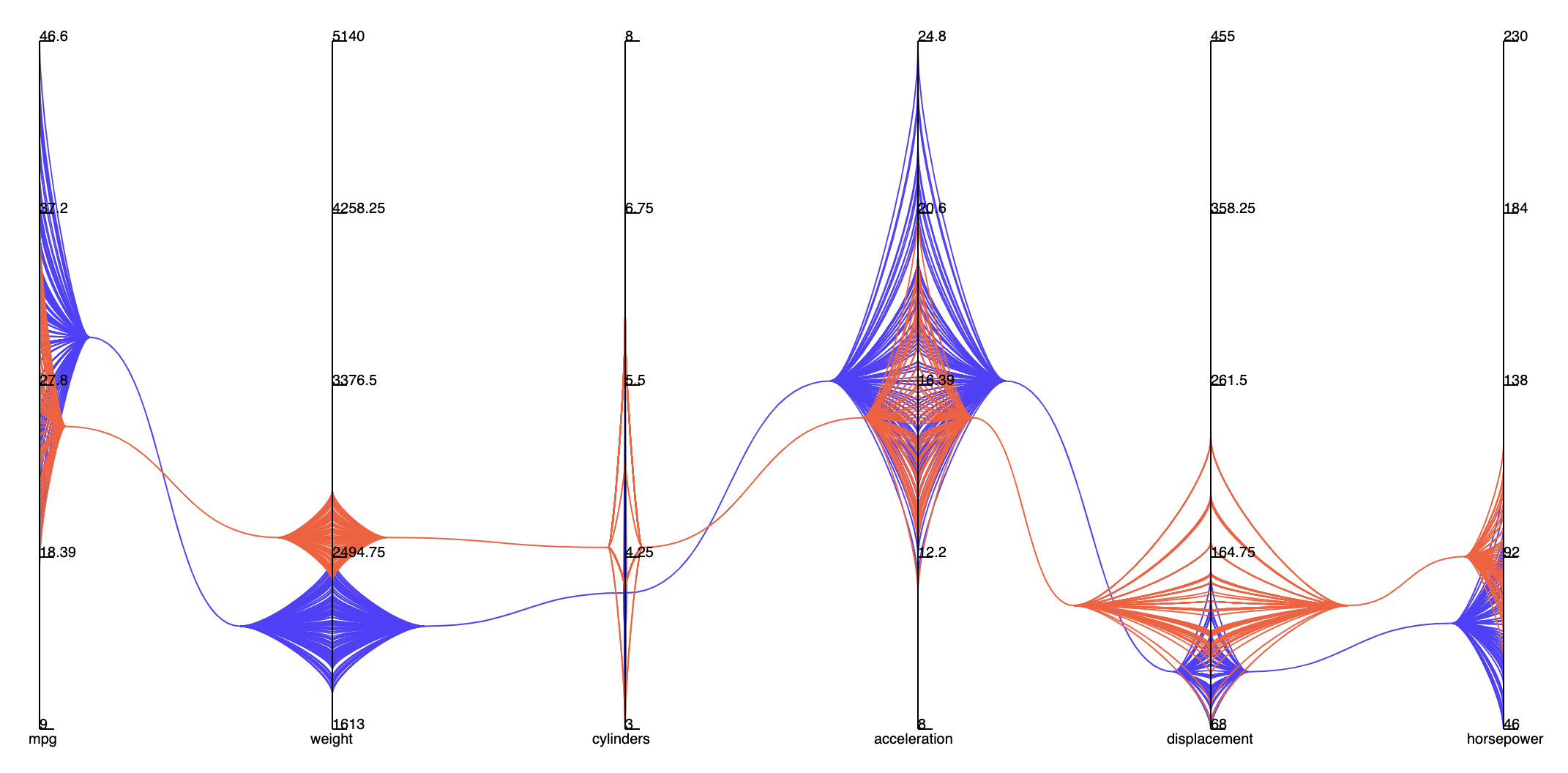}}
	\hfill
	\subfloat[The car data set visualized in our confluent-drawing PCP\label{correlationCD}]{%
		\includegraphics[width=0.9\linewidth]{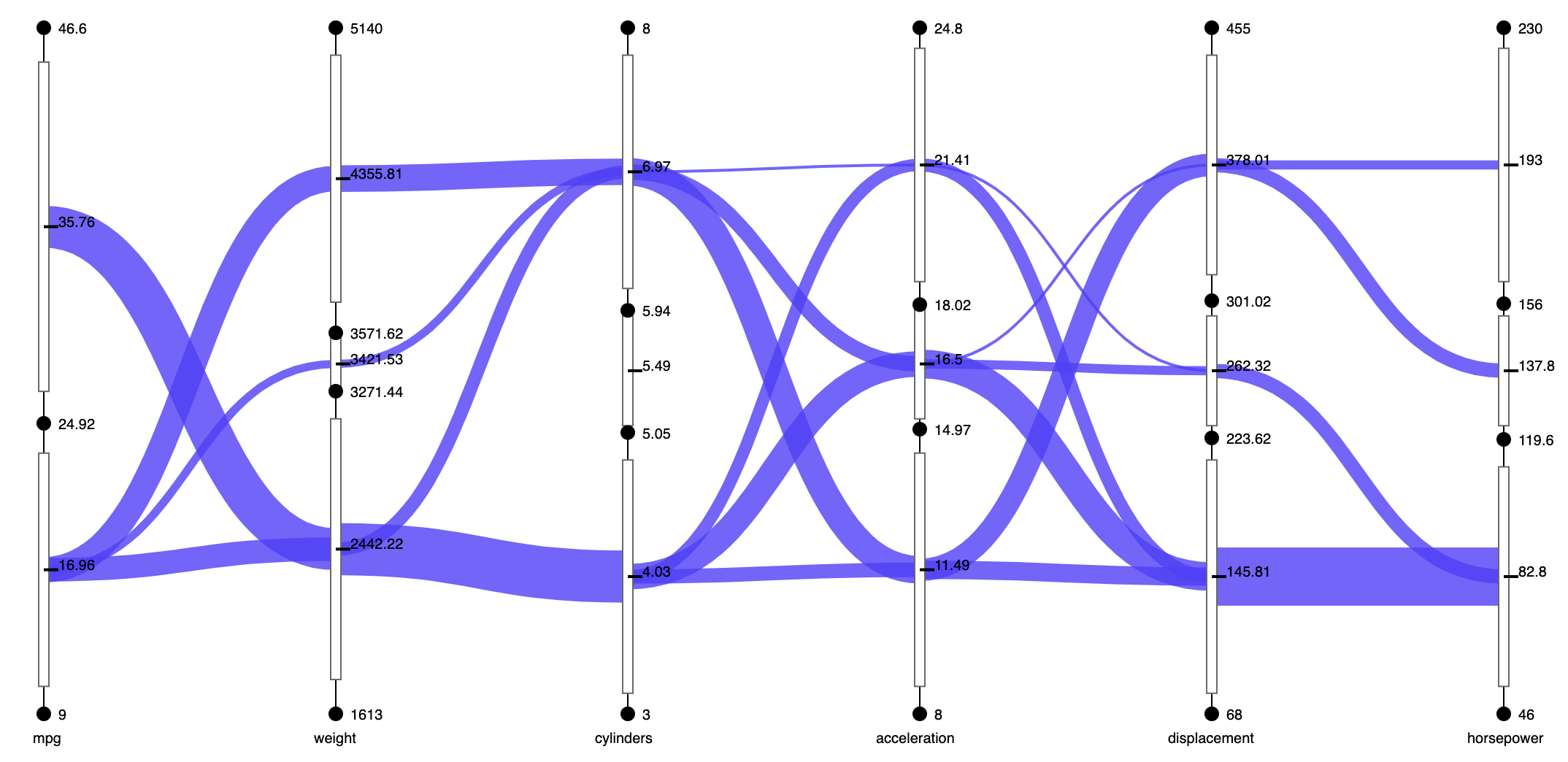}}
	\caption{The comparison of different PCP layouts with the car data set. Visualization used in the user study for the correlation tasks.}
	\label{fig:correlationVis} 
\end{figure}

To evaluate the effectiveness of our method, we conducted a comparative user study of our method with do Amor Divino Lima et al.'s edge-bundling PCP~\cite{Do2018}. Two typical tasks, correlation estimation and subset estimation were used in the user study, in which accuracy was used as the sole dependent measure. The study was performed using static images. \autoref{correlationEB} and \autoref{correlationCD} were used in the correlation estimation. \autoref{room2} and \autoref{room3_na} were used in the clusters estimation. The two tasks are described as follows:
\begin{description}
	\item [Correlation Estimation.] Participants are asked to estimate the correlation of two variables on a discrete scale from -1 to +1 with the interval of 0.1. +1 denotes positive correlation. -1 denotes negative correlation. 0 denotes that there is no correlation. Example task: Estimate the correlation coefficient of displacement and horsepower in \autoref{correlationEB} and \autoref{correlationCD}.
	\item [Subset Estimation.] Participants are asked to estimate the number of subsets over several axes. A subset is a set of clusters that have common data items on different axes, which are shown as a common path over the axes. For example, between the axes of temperature and light, there are 4 subsets in \autoref{room3_na} and 3 subsets in \autoref{room2}.  
\end{description}


16 students and researchers at a local university participated the user study. Each task was carefully explained to the participants. They divided into two groups. Both groups did the two tasks with the two methods but in different order. One group first did the tasks with our method. Another group first did the tasks with the edge-bundling PCP. In the questionnaire, for each method, there are 5 questions of correlation estimation, and 4 questions of subset estimation. To preprocess the data, we removed the answers of participants who did not complete the tasks, and the answers with an error larger than the mean error plus 2 times standard deviations. For the correlation estimation task, 1 participant's answers are removed (5 out of 80 answers). For the subset estimation task, 10 out of 64 answers are removed.

For each method, to assess the accuracy of the two tasks, we calculated the mean error of all the answers in each task. For each task, the error of an answer is computed as follows:
\begin{description}
	\item [Correlation Estimation.] The error is the absolute difference between the participant's answer and the ground-truth correlation (Pearson correlation coefficient).
	\item [Subset Estimation.] The error is the absolute difference between the participant's answer and the ground-truth subset number.
\end{description}

\begin{figure} [tb]
	\centering
	\subfloat[The mean error of the correlation estimation task.\label{correlation_error}]{%
		\includegraphics[width=0.8\linewidth]{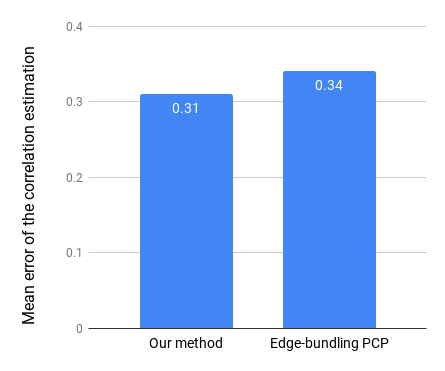}}
	\hfill
	\subfloat[The mean error of the subset estimation task. \label{subset_error}]{%
		\includegraphics[width=0.8\linewidth]{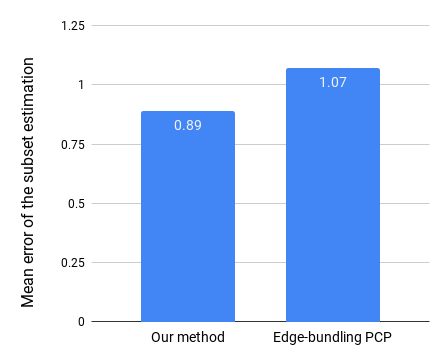}}
	\caption{The comparison of the mean errors.}
	\label{fig:mean_error} 
\end{figure}

\autoref{fig:mean_error} shows the mean errors of the two tasks for each method. According to~\autoref{correlation_error} and~\autoref{subset_error}, in both of the correlation estimation and subset estimation tasks, our method has better performance (smaller mean errors) than do Amor Divino Lima et al.'s edge-bundling PCP. This is because our method eliminates overlaps on clusters and edges, which reveals all information of the data. In contrast, do Amor Divino Lima et al.'s edge-bundling PCP creates the obvious overlaps on clusters on each axis, which is shown in~\autoref{room2} and~\autoref{correlationEB}. \autoref{room2} also shows the overlaps on the edges between the axes of light and occupancy, which is the reason why do Amor Divino Lima et al.'s edge-bundling PCP has a larger mean error for the subset estimation task. Moreover, do Amor Divino Lima et al.'s edge-bundling PCP eliminates much data points in the data set by the clustering method they used, which may lead to the misunderstanding of the whole data set.


Based on the experiments and user study, we can see that our method has clear advantages for the web-based interactive visual analytics of large multi-dimensional data. With its scalability, it supports real-time interactions on millions of data items without hardware acceleration in a normal web browser. For the typical tasks of PCP, it has better performances than the compared edge-bundling PCP. In addition, based on the proposed clustering method, it can display the major clusters and outliers without the assistance of the color and opacity channels.

\section{Conclusion}\label{section:future}

In this paper, we proposed a confluent-drawing PCP for the web-based visual analytics of large multi-dimensional data. Based on the proposed approach, we implement a web-based visual analytics application. We evaluated the scalability and effectiveness of the proposed approach through the experiments and user study of comparing it with another edge-bundling PCP. Results show that the proposed approach significantly enhances the web-based interactive visual analytics of large multi-dimensional data.

The work in this paper can be further improved in several ways. To improve the visualization, the color and opacity perceptual channels could be investigated to improve its performance. For example, use color to highlight the related subsets (common paths) when hovering over one bundling.


\bibliographystyle{abbrv-doi}

\bibliography{bibliography}
\end{document}